\documentclass[11pt,a4paper]{article}
\usepackage{hyperref}
\usepackage[hyperref]{naaclhlt2018}
\usepackage{times}
\usepackage{latexsym}

\usepackage{url}

\usepackage{graphicx}

\definecolor{mention}{HTML}{7FC97F}
\definecolor{attention}{HTML}{47494C}

\usepackage{amsmath}
\usepackage{bm}

\usepackage{amssymb}

\usepackage{amsopn}
\DeclareMathOperator*{\argmax}{argmax}

\usepackage{cleveref}
\crefname{section}{§}{§§}
\Crefname{section}{§}{§§}
\Crefname{figure}{Figure}{}
\Crefname{algorithm}{Algorithm}{}
\Crefname{equation}{Eq.}{}

\usepackage{siunitx}
\newcommand{\superscript}[1]{\ensuremath{^{\textrm{#1}}}}

\usepackage{booktabs}

\usepackage{multirow}

\usepackage{array}
\newcolumntype{L}[1]{>{\raggedright\let\newline\\\arraybackslash\hspace{0pt}}m{#1}}
\newcolumntype{C}[1]{>{\centering\let\newline\\\arraybackslash\hspace{0pt}}m{#1}}
\newcolumntype{R}[1]{>{\raggedleft\let\newline\\\arraybackslash\hspace{0pt}}m{#1}}

\newlength{\Width}%

\newcommand{\MyColorBox}[2][red]%
{%
    \settowidth{\Width}{#2}%
    \colorbox{#1}%
    {%
        \raisebox{-\DepthReference}%
        {%
                \parbox[b][\HeightReference+\DepthReference][c]{\Width}{\centering#2}%
        }%
    }%
}

\aclfinalcopy 
\def\aclpaperid{} 

\title{Fine-grained Entity Typing through  Increased Discourse Context\\ and Adaptive Classification Thresholds}

\author{Sheng Zhang\\
  Johns Hopkins University\\
  {\tt zsheng2@jhu.edu} \\\And
  Kevin Duh\\
  Johns Hopkins University\\
  {\tt kevinduh@cs.jhu.edu} \\\And
  Benjamin Van Durme\\
  Johns Hopkins University\\
  {\tt vandurme@cs.jhu.edu} \\}

\date{}

\begin{document}
\maketitle
\begin{abstract}
   Fine-grained entity typing is the task of assigning fine-grained semantic types to 
entity mentions. 
We propose a neural architecture
    which learns a distributional semantic  representation that leverages a greater amount of semantic context -- both document and sentence level information -- than prior work.  We find that additional context improves performance, with further improvements gained by utilizing adaptive classification thresholds.
Experiments show that
our approach without reliance on hand-crafted features achieves the state-of-the-art results on three benchmark datasets.
\end{abstract}

\section{Introduction}

\begin{figure}[t]
\centering
\includegraphics[width=0.49\textwidth]{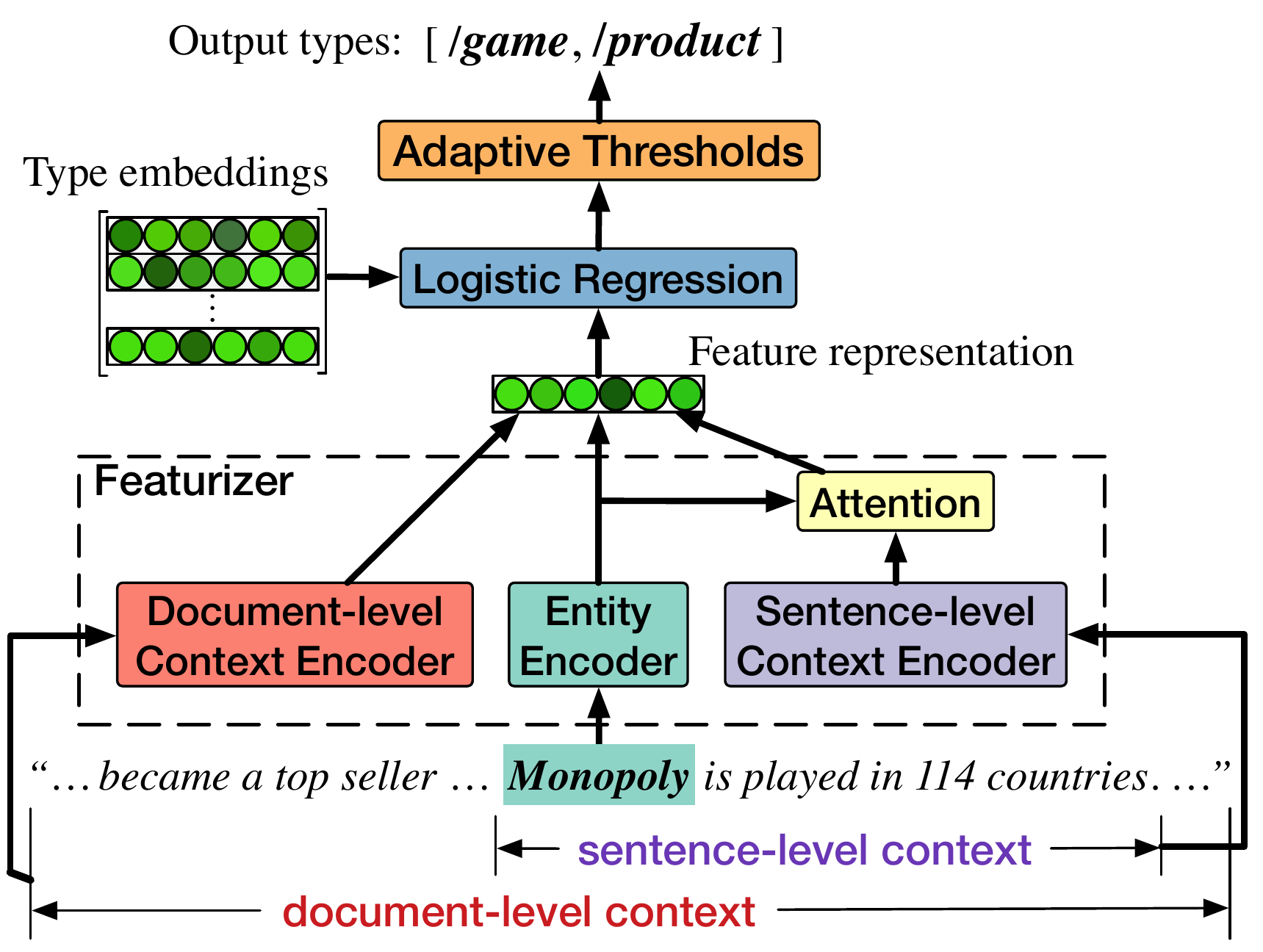}
\caption{Neural architecture for predicting the
    types of entity mention ``{\em Monopoly}" in the text
    ``{\em... became a top seller ... Monopoly is played
    in 114 countries. ...}''. Part of document-level context is omitted.
    \label{fig:arch}}
\end{figure}

Named entity typing is the task of detecting the type (e.g., \textit{person}, 
\textit{location}, or \textit{organization}) of a named entity in natural language
text. Entity type information has shown to be useful in 
natural language tasks such as question answering~\cite{Lee2006}, knowledge-base
population~\cite{carlson2010coupled,DBLP:conf/aaai/MitchellCHTBCMG15}, and co-reference resolution~\cite{recasens-demarneffe-potts:2013:NAACL-HLT}. 
Motivated by its application to downstream tasks,
recent work on entity typing has moved beyond standard coarse types
towards finer-grained semantic types with richer ontologies~\cite{Lee2006,Ling2012,yosef-EtAl:2012:POSTERS,gillick2014context,delcorro-EtAl:2015:EMNLP}. Rather than assuming an entity can be uniquely categorized into a single type, the task has been approached as a multi-label classification
problem: e.g., in ``{\em... became a top seller ... Monopoly is played
in 114 countries. ...}'' (\Cref{fig:arch}), ``{\em Monopoly}'' is considered
both a \emph{game} as well as a \emph{product}.

The state-of-the-art approach~\cite{shimaoka-EtAl:2017:EACLlong}
for fine-grained entity typing employs an attentive neural
architecture to learn representations of the entity mention as well as its context.
These representations are then combined with hand-crafted features
(e.g., lexical and syntactic features),
and fed into a linear classifier with a fixed threshold. 
While this approach outperforms previous approaches 
which only use sparse binary features~\cite{Ling2012,gillick2014context}
or distributed representations~\cite{yogatama-gillick-lazic:2015:ACL-IJCNLP},
it has a few drawbacks:
(1) the representations of left and right contexts are learnt independently,
ignoring their mutual connection;
(2) the attention on context is computed solely upon the context,
considering no alignment to the entity;
(3) document-level contexts which could be useful in classification are not
exploited;
and (4) hand-crafted features heavily rely on system or human annotations.

To overcome these drawbacks, we propose a neural architecture (\Cref{fig:arch})
which learns more context-aware representations by using a better attention
mechanism and taking advantage of semantic discourse information available in both the document as well as sentence-level contexts.  Further, we find that adaptive classification thresholds leads to further improvements. Experiments demonstrate that our approach, without any reliance on hand-crafted features, outperforms prior work on three benchmark datasets.

\section{Model}
Fine-grained entity typing is considered a multi-label
classification problem: Each entity $e$ in the text $x$ is assigned a set
of types $T^*$ drawn from the fine-grained type set $\mathcal{T}$.
The goal of this task is to predict, given entity $e$ and its context $x$,
the assignment of types to the entity.
This assignment can be represented by a binary vector $y\in\{1,0\}^{|\mathcal{T}|}$
where $|\mathcal{T}|$ is the size of $\mathcal{T}$.
$y_t=1$ iff the entity is assigned type $t\in \mathcal{T}$.

\subsection{General Model}
Given a type embedding vector $w_t$ and a featurizer $\varphi$ that takes entity $e$ and its context $x$,
we employ the logistic regression (as shown in \Cref{fig:arch})
to model the probability of $e$ assigned $t$ (i.e., $y_t=1$)
\begin{equation} \label{eq:prob}
    P(y_t=1) = \frac{1}{1 + \exp{(-w^{\intercal}_t\varphi(e, x))}},
\end{equation}
and we seek to learn a type embedding matrix $W =[w_1, \ldots, w_{|\mathcal{T}|}]$ and 
a featurizer $\varphi$ such that
\begin{equation} \label{eq:loss}
    T^* = \argmax_{T}\prod_{t\in T}P(y_t=1)\cdot\prod_{t\notin T}P(y_t=0).
\end{equation}

At inference, the predicted type set $\hat{T}$ assigned to entity $e$
is carried out by
\begin{equation}
    \hat{T} = \big\{t\in\mathcal{T}: P(y_t=1) \geq r_t \big\},
\end{equation}
with $r_t$  the threshold for predicting $e$ has type $t$.

\subsection{Featurizer}
As shown in \Cref{fig:arch}, featurizer $\varphi$ in our model contains three encoders
which encode entity $e$ and its context $x$ into feature vectors,
and we consider both {\em sentence-level} context $x_s$ and {\em document-level}
context $x_d$
in contrast to prior work which only takes {\em sentence-level} context
\cite{gillick2014context,shimaoka-EtAl:2017:EACLlong}.
\footnote{Document-level context has also been exploited in
\newcite{yaghoobzadeh-schutze:2015:EMNLP,yang-EtAl:2016:N16-13,karn-waltinger-schutze:2017:EACLshort,gupta-singh-roth:2017:EMNLP2017}.}

The output of featurizer $\varphi$ is the concatenation of these feature vectors:
\begin{align}
    \varphi(e, x) = \left [
        \begin{gathered}
            f(e)\\
            g_s(x_s, e)\\
            g_d(x_d)
    \end{gathered}
\right ].
\end{align}
We define the computation of these feature vectors in the followings.

\noindent{\bf Entity Encoder:}
The entity encoder $f$ computes the average of all the embeddings of tokens in entity $e$.

\noindent{\bf Sentence-level Context Encoder:}
The encoder $g_s$ for sentence-level context $x_s$ employs a single bi-directional
RNN to encode $x_s$. 
Formally, let the tokens in $x_s$ be $x^1_s,\ldots,x^n_s$.
The hidden state $h_i$ for token $x^i_s$ is a concatenation of a left-to-right
hidden state $\overrightarrow{h_i}$ and a right-to-left hidden state
$\overleftarrow{h_i}$,
\begin{align}
    h_i = \left [
        \begin{aligned}
        \overrightarrow{h}_i\\
        \overleftarrow{h}_i
    \end{aligned}
\right ] = \left [
    \begin{aligned}
        \overrightarrow{f}(x^i_s, \overrightarrow{h}_{i-1})\\
        \overleftarrow{f}(x^i_s, \overleftarrow{h}_{i+1})
    \end{aligned}
\right],
\end{align}
where $\overrightarrow{f}$ and $\overleftarrow{f}$ are $L$-layer stacked
LSTMs units~\cite{hochreiter1997long}. 
This is different from \newcite{shimaoka-EtAl:2017:EACLlong} who use two separate
bi-directional RNNs for context on each side of the entity mention.

\noindent{\bf Attention:}
The feature representation for $x_s$ is a weighted sum of the hidden states:
$g_s(x_s, e) = \sum_{i=1}^na_ih_i$, where $a_i$ is the attention to hidden
state $h_i$.
We employ the dot-product attention~\cite{luong-pham-manning:2015:EMNLP}.
It computes attention based on the alignment between the entity and its context:

\begin{align}
    a_i = \frac{\exp{(h^{\intercal}_iW_af(e))}}{\sum_{j=1}^n\exp{(h^{\intercal}_jW_af(e))}},
\end{align}
where $W_a$ is the weight matrix.
The dot-product attention differs from the self attention~\cite{shimaoka-EtAl:2017:EACLlong} which only considers the context.

\noindent{\bf Document-level Context Encoder:}
The encoder $g_d$ for document-level context $x_d$ is a multi-layer
perceptron:
\begin{align}
    g_d(x_d) = \textrm{relu}(W_{d_1}\textrm{tanh}(W_{d_2}\textrm{DM}(x_d))),
\end{align}
where DM is a pretrained distributed memory model
\cite{le2014distributed}
which converts the document-level context into a distributed representation.
$W_{d_1}$ and $W_{d_2}$ are weight matrices.

\subsection{Adaptive Thresholds}
In prior work, a fixed
threshold ($r_t=0.5$) is used for classification of all types~\cite{Ling2012,shimaoka-EtAl:2017:EACLlong}.
We instead assign a different threshold to each type that is optimized to maximize
the overall strict $F_1$ on the dev set.  We show the definition of strict $F_1$ in Section\Cref{subsec:metrics}. 

\section{Experiments}
We conduct experiments on three publicly available datasets.\footnote{
We made the source code and data publicly available at \url{https://github.com/sheng-z/figet}.}
\Cref{tab:stat} shows the statistics of these datasets.

\noindent{\bf OntoNotes:}
\newcite{gillick2014context} sampled sentences from OntoNotes
\cite{weischedel2011ontonotes} and annotated entities in these sentences
using 89 types.
We use the same train/dev/test splits in \newcite{shimaoka-EtAl:2017:EACLlong}.
Document-level contexts are retrieved
from the original OntoNotes corpus.

\noindent{\bf BBN:} \newcite{weischedel2005bbn} annotated entities in Wall Street Journal
using 93 types.
We use the train/test splits in~\newcite{Ren:2016:LNR:2939672.2939822}
and randomly hold out 2,000 pairs for dev.
Document contexts are retrieved from the original corpus.

\noindent{\bf FIGER:} \newcite{Ling2012} sampled sentences from ~780k Wikipedia
articles and 434 news reports to form the train and test data respectively,
and annotated entities using 113 types.
The splits we use are the same in \newcite{shimaoka-EtAl:2017:EACLlong}.

\begin{table}[h]
\centering
\small
\label{tab:splits}
\begin{tabular}{@{}lcccc@{}}
\toprule
                   & \textbf{Train} & \textbf{Dev} & \textbf{Test} & \textbf{Types}\\ \midrule
    \textbf{OntoNotes} & 251,039        & 2,202        & 8,963       & 89  \\
    \textbf{BBN}       & 84,078         & 2,000       & 13,766      & 93  \\ 
    \textbf{FIGER}     & 2,000,000      & 10,000       & 563        & 113   \\ \bottomrule
\end{tabular}
\caption{Statistics of the datasets.\label{tab:stat}}
\end{table}


\subsection{Metrics} \label{subsec:metrics}
We adopt the metrics used in~\newcite{Ling2012} where 
results are evaluated via strict, loose macro, loose micro $F_1$ scores.
For the $i$-th instance, let the predicted type set be $\hat{T_i}$, and the
reference type set $T_i$. The precision ($P$) and recall ($R$) for each metric are computed as follow.

\noindent{\bf Strict}:
\begin{align*}
P = R = \frac{1}{N}\sum_{i=1}^N\delta(\hat{T_i}=T_i)
\end{align*}
\noindent{\bf Loose Macro}:
\begin{align*}
    P = \frac{1}{N}\sum_{i=1}^N\frac{|\hat{T_i}\cap T_i|}{|\hat{T_i}|} \\
    R = \frac{1}{N}\sum_{i=1}^N\frac{|\hat{T_i}\cap T_i|}{|T_i|}
\end{align*}
\noindent{\bf Loose Micro}:
\begin{align*}
    P = \frac{\sum_{i=1}^N|\hat{T_i}\cap T_i|}{\sum_{i=1}^N|\hat{T_i}|} \\
    R = \frac{\sum_{i=1}^N|\hat{T_i}\cap T_i|}{\sum_{i=1}^N|T_i|}
\end{align*}

\subsection{Hyperparameters}
\label{sec:hyper}
We use open-source GloVe vectors~\cite{pennington2014glove} trained on Common
Crawl 840B with 300 dimensions
to initialize word embeddings used in all encoders.
All weight parameters are sampled from $\mathcal{U}(-0.01, 0.01)$. 
The encoder for sentence-level context is a 2-layer bi-directional RNN with
200 hidden units. The DM output size is 50.
Sizes of $W_a$, $W_{d_1}$ and $W_{d_2}$ are $200\times 300$, $70\times 50$,
and $50\times 70$ respectively. Adam optimizer~\cite{kingma2014adam} and mini-batch gradient is used
for optimization. Batch size is 200.
Dropout (rate=0.5) is applied to three feature functions. To avoid overfitting,
we choose models which yield the best strict $F_1$ on dev sets.

\subsection{Results}
\label{subsec:results}
We compare experimental results of our approach with previous
approaches\footnote{For PLE \cite{Ren:2016:LNR:2939672.2939822}, we were unable to replicate the performance benefits 
reported in their work, so we report the results after running their codebase.},
and study contribution of our base model architecture, document-level contexts and adaptive thresholds via ablation.
To ensure our findings are reliable, we run each experiment twice and report the average performance.

Overall, our approach significantly increases the state-of-the-art macro $F_1$ on both OntoNotes and BBN datasets.

\begin{table*}[htp]
\centering
\small
    \begin{tabular}{@{}cL{0.5\textwidth}L{0.18\textwidth}L{0.18\textwidth}@{}}
\toprule
        \textbf{ID} & \textbf{Sentence} & \textbf{Gold} & \textbf{Prediction}             \\ \midrule
        A & \textit{... Canada's declining crude output, combined with ...}
        \textit{\colorbox{attention!8!}{\rule[0pt]{0pt}{6pt}will} \colorbox{attention!55!}{\rule[0pt]{0pt}{6pt}help} \colorbox{attention!34!}{\rule[0pt]{0pt}{6pt}intensify} \colorbox{attention!2!}{\rule[0pt]{0pt}{6pt}U.S.} reliance on \colorbox{mention}{\rule[0pt]{0pt}{6pt}oil} from overseas. ...}
        & /other & /other \newline \textcolor{black!50}{/other/health} \newline \textcolor{black!50}{/other/health/treatment} \\ \midrule
        B & \textit{Bozell joins Backer Spielvogel Bates and \colorbox{mention}{\rule[0pt]{0pt}{6pt}Ogilvy Group} as \colorbox{attention!14!}{\rule[0pt]{0pt}{6pt}U.S.} \colorbox{attention!52!}{\rule[0pt]{0pt}{6pt}agencies} \colorbox{attention!23!}{\rule[0pt]{0pt}{6pt}with} \colorbox{attention!9!}{\rule[0pt]{0pt}{6pt}interests} in Korean agencies.} & /organization \newline /organization/company & /organization \newline /organization/company \\ 
\bottomrule
\end{tabular}
\caption{Examples showing the improvement brought by document-level contexts and dot-product attention.\\
    {Entities are shown in the green box. The gray boxes visualize attention weights (darkness) on context tokens.}}
\label{tab:cases}
\end{table*}

\begin{table}[htp]
\setlength{\tabcolsep}{3pt}
\centering
\small
\sisetup{
            detect-all,
            table-number-alignment = center,
            table-figures-integer = 2,
            table-figures-decimal = 2,
            table-space-text-post = {\superscript{*}},
}
\begin{tabular}{@{}L{0.28\textwidth}@{\hskip 1pt}ccS@{}}
\toprule
                         \textbf{Approach}& \textbf{Strict} & \textbf{Macro} & \textbf{Micro} \\ \midrule
    \textsc{Binary}\cite{gillick2014context}         & N/A               & N/A              & 70.01         \\
\textsc{Kwsabie}\cite{yogatama-gillick-lazic:2015:ACL-IJCNLP}       & N/A               & N/A              & 72.98          \\ \midrule\midrule
\textsc{PLE}\cite{Ren:2016:LNR:2939672.2939822}             & 51.61           & 67.39          & 62.38          \\
\newcite{C16-1017} & 49.30 & 68.23 & 61.27 \\
\textsc{AFET}\cite{ren-EtAl:2016:EMNLP2016}             & 55.10           & 71.10          & 64.70          \\
\textsc{Fnet}\cite{abhishek-anand-awekar:2017:EACLlong} & 52.20 & 68.50 & 63.30 \\
\textsc{Neural}\cite{shimaoka-EtAl:2017:EACLlong}         & 51.74           & 70.98          & 64.91          \\ 
    \hspace{0.3cm}w/o Hand-crafted features          & 47.15 & 65.53 & 58.25\\ \midrule
\textsc{Our Approach} &       \textbf{55.52}          &      \textbf{73.33}          &       \textbf{67.61}         \\
\hspace{0.3cm}w/o Adaptive thresholds &  53.49               &        73.11        &      66.78          \\
\hspace{0.3cm}w/o Document-level contexts&  53.17               &        72.14        &      66.51          \\
    \hspace{0.3cm}w/~~ Hand-crafted features          & 54.40 & 73.13 & 66.89\\ 

\bottomrule
\end{tabular}
\caption{Results on the OntoNotes dataset. }
\label{tab:ontonotes}
\end{table}

On OntoNotes (\Cref{tab:ontonotes}), our approach improves the state of the art across all three metrics.
Note that (1) without adaptive thresholds or document-level contexts, our approach
still outperforms other approaches on macro $F_1$ and micro $F_1$;
(2) adding hand-crafted features~\cite{shimaoka-EtAl:2017:EACLlong} does not improve the performance.
This indicates the benefits of our proposed model architecture for learning fine-grained entity typing,
which is discussed in detail in Section\Cref{sec:ana}; 
and (3) B\textsc{inary} and K\textsc{wasibie} were trained on a different dataset, so their results are not directly comparable.

\begin{table}[htp]
\setlength{\tabcolsep}{3pt}
\centering
\small
\begin{tabular}{@{}L{0.28\textwidth}@{\hskip 1pt}ccc@{}}
\toprule
                        \textbf{Approach} & \textbf{Strict} & \textbf{Macro} & \textbf{Micro} \\ \midrule
\textsc{PLE}\cite{Ren:2016:LNR:2939672.2939822}            & 49.44           & 68.75          & 64.54          \\ 
\newcite{C16-1017} & \textbf{70.43} & 75.78 & 76.50 \\
    \textsc{AFET}\cite{ren-EtAl:2016:EMNLP2016}             & 67.00           & 72.70          & 73.50          \\ 
        \textsc{Fnet}\cite{abhishek-anand-awekar:2017:EACLlong} & 60.40 & 74.10 & 75.70 \\
\midrule
    \textsc{Our Approach} &       60.87          &      \textbf{77.75}          &       \textbf{76.94}         \\
\hspace{0.3cm}w/o Adaptive thresholds &  58.47               &        75.84        &      75.03          \\
\hspace{0.3cm}w/o Document-level contexts&  58.12               &        75.65        &      75.11          \\
\bottomrule
\end{tabular}
\caption{Results on the BBN dataset.}
\label{tab:bbn}
\end{table}

On BBN (\Cref{tab:bbn}), while \newcite{C16-1017}'s label embedding algorithm
holds
the best strict $F_1$, our approach notably improves both macro $F_1$ and micro $F_1$.\footnote{
    Integrating label embedding into our proposed approach is an avenue for future work.}
The performance drops to a competitive level with other approaches if adaptive thresholds or document-level contexts are removed.

\begin{table}[htp]
\setlength{\tabcolsep}{3pt}
\centering
\small
\begin{tabular}{@{}L{0.28\textwidth}@{\hskip 1pt}ccc@{}}
\toprule
                         \textbf{Approach}& \textbf{Strict} & \textbf{Macro} & \textbf{Micro} \\ \midrule
\textsc{Kwsabie}\cite{yogatama-gillick-lazic:2015:ACL-IJCNLP}         & N/A               & N/A              & 72.25           \\ 
\text{Attentive}\cite{shimaoka-EtAl:2016:W16-13}  & 58.97   & 77.96   & 74.94   \\
    \textsc{Fnet}\cite{abhishek-anand-awekar:2017:EACLlong} & 65.80  & 81.20  & 77.40   \\
\midrule\midrule
    \newcite{Ling2012}          &  52.30              &  69.90             & 69.30          \\
\textsc{PLE}\cite{Ren:2016:LNR:2939672.2939822}            & 49.44           & 68.75          & 64.54          \\ 
\newcite{C16-1017} & 53.54 & 68.06 & 66.53 \\
    \textsc{AFET}\cite{ren-EtAl:2016:EMNLP2016}             & 53.30           & 69.30          & 66.40          \\ 
    \textsc{Neural}\cite{shimaoka-EtAl:2017:EACLlong}          & 59.68 & \textbf{78.97} & 75.36\\
    \hspace{0.3cm}w/o Hand-crafted features          & 54.53 & 74.76 & 71.58\\ \midrule
    \textsc{Our Approach} &       \textbf{60.23}         &      78.67          &       \textbf{75.52}         \\
    \hspace{0.3cm}w/o Adaptive thresholds &  60.05 & 78.50        &      75.39          \\
    \hspace{0.3cm}w/~~ Hand-crafted features          & 60.11 & 78.54 & 75.33 \\ 
\bottomrule
\end{tabular}
\caption{Results on the FIGER dataset.}
\label{tab:figer}
\end{table}

On FIGER (\Cref{tab:figer}) where no document-level context is currently available, our proposed approach
still achieves the state-of-the-art strict and micro $F_1$.
If compared with the ablation variant of the \textsc{\small Neural}
approach, i.e., w/o hand-crafted features, our approach gains significant
improvement.
We notice that removing adaptive thresholds only causes a small performance
drop;
this is likely because 
the train and test splits of FIGER are from different sources, and
adaptive thresholds are not generalized well enough to the test data.
K\textsc{wasibie}, Attentive and F\textsc{net} were trained on a different dataset, so their results are not directly comparable.


\subsection{Analysis}
\label{sec:ana}
\Cref{tab:cases} shows examples illustrating the benefits brought by our proposed approach.
Example A illustrates that sentence-level context sometimes is not informative enough,
and attention, though already placed on the head verbs, can be misleading.
Including document-level context (i.e., ``\textit{Canada's declining crude output}'' in this case)
helps preclude wrong predictions
(i.e., /other/health and /other/health/treatment). Example B shows that the semantic patterns learnt by our attention mechanism
help make the correct prediction. As we observe in \Cref{tab:ontonotes} and \Cref{tab:figer},
adding hand-crafted features to our approach does not improve the results.
One possible explanation is that hand-crafted features are mostly about
syntactic-head or topic information, and such information
are already covered by our attention mechanism and document-level contexts
as shown in \Cref{tab:cases}.
Compared to hand-crafted features that heavily rely on system or human annotations,
attention mechanism requires significantly less supervision, and document-level or paragraph-level 
contexts are much easier to get.

Through experiments, we observe no improvement by encoding type hierarchical information~\cite{shimaoka-EtAl:2017:EACLlong}.\footnote{The type embedding matrix $W$
for the logistic regression is replaced by the product of a learnt weight
matrix $V$ and the constant sparse binary matrix $S$ which encodes type hierarchical information.}
To explain this, we compute cosine similarity between each pair of fine-grained types
based on the type embeddings learned by our model, i.e., $w_t$ in \Cref{eq:prob}. 
\Cref{tab:type-sim} shows several types and their closest types:
these types do not always share coarse-grained types with their
closest types,
but
they often co-occur in the same context.


\begin{table}[htp]
\setlength{\tabcolsep}{3pt}
\centering
\small
\begin{tabular}{@{}ll@{}}
\toprule
\multicolumn{1}{c}{\textbf{Type}}          & \multicolumn{1}{c}{\textbf{Closest Types}} \\ \midrule
\multirow{2}{*}{/other/event/accident}& /location/transit/railway\\
& /location/transit/bridge \\ \midrule
\multirow{2}{*}{/person/artist/music}& /organization/music \\
& /person/artist/director \\ \midrule
\multirow{2}{*}{/other/product/mobile\_phone}& /location/transit/railway \\
& /other/product/computer \\ \midrule
\multirow{2}{*}{/other/event/sports\_event}& /location/transit/railway \\
& /other/event \\ \midrule
\multirow{2}{*}{/other/product/car}& /organization/transit \\
& /other/product \\ \midrule
\end{tabular}
    \caption{Type similarity.}
    \label{tab:type-sim}
\end{table}

\section{Conclusion}
We propose a new approach for fine-grained entity typing. 
The contributions are: 
(1) we propose a neural architecture which learns a distributional semantic 
representation that leverage both document and sentence level information,
(2) we find that context increased with document-level information improves 
performance,  and (3) we utilize adaptive classification thresholds to further boost the performance.
Experiments show our approach achieves new state-of-the-art results on three
benchmarks.

\section*{Acknowledgments}

This work was supported in part by the JHU Human Language Technology
Center of Excellence (HLTCOE), and DARPA
LORELEI. The U.S. Government is authorized to reproduce and distribute
reprints for Governmental purposes. The views and conclusions
contained in this publication are those of the authors and should not
be interpreted as representing official policies or endorsements of
DARPA or the U.S. Government.

\bibliography{naaclhlt2018}
\bibliographystyle{acl_natbib}

%
%

\end{document}